\pdfoutput=1
\documentclass{article}
\usepackage{iclr2027_conference,times}

\usepackage{amsmath,amsfonts,bm}

\def\eqref#1{equation~\ref{#1}}

\def\1{\bm{1}}

\DeclareMathAlphabet{\mathsfit}{\encodingdefault}{\sfdefault}{m}{sl}
\SetMathAlphabet{\mathsfit}{bold}{\encodingdefault}{\sfdefault}{bx}{n}

\usepackage{booktabs}
\usepackage{flafter}
\usepackage{graphicx}
\usepackage{multirow}
\usepackage{placeins}
\usepackage{tikz}
\usetikzlibrary{arrows.meta,positioning}
\usepackage{hyperref}
\usepackage{url}
\usepackage{enumitem}
\usepackage{bbm}
\usepackage{algorithm2e}
\usepackage{adjustbox}

\newcommand{\myparagraph}[1]{{\setlength{\parskip}{0.3em} \noindent \textbf {#1}}}

\definecolor{darkgreen}{rgb}{0.0, 0.4, 0.2}

\newcommand{\projecturl}{https://wrecklong.github.io/CtrlCache/}

\title{CtrlCache: Accelerating Interactive Video World Models with Control-Aware Caching}

\author{Shangye Song$^{1}$ \quad Dong Gong$^{2}$ \quad Hong Jia$^{1}$ \quad Yun Sing Koh$^{1}$ \quad Xinyu Zhang$^{1}$\thanks{Corresponding author.} \\
$^{1}$University of Auckland~~ $^{2}$ UNSW Sydney \\
\texttt{\{shangye.song,hong.jia,y.koh,xinyu.zhang\}@auckland.ac.nz} \\
\texttt{dong.gong@unsw.edu.au} \\
\textbf{Project page:} \url{\projecturl}
}

\iclrfinalcopy

\begin{document}

\maketitle
\fancyhead{}

\begin{abstract}

Interactive video world models need to generate each video chunk efficiently while responding faithfully to user controls.
Many systems use chunk-wise
autoregressive generation with few-step denoising, but each chunk still
requires several costly denoising iterations.
Training-free caching can reduce this cost, yet existing policies make reuse decisions primarily from model-internal denoising dynamics and do not
explicitly account for control transitions.
Actually, interactive
generation explicitly exposes a signal they do not use: the controls for a chunk arrive before it is denoised, so a schedule derived from them costs no forward pass.
To this end, we analyze adjacent chunks under different control regimes and
find that structural similarity drops around action changes, while low-frequency
structure remains more persistent than high-frequency detail.
Motivated by these observations, we propose \textbf{CtrlCache}, a training-free control-aware caching framework that adapts computation to the
current control sequence.
Specifically, the action-aware scheduling and refresh policy detects action
changes across and within chunks, and labels each chunk as initial, transition,
turning, or steady state.
At one selected interior
denoising step, initial and transition chunks retain full computation, while turning
and steady chunks reuse the transformer residual from the most recent fully computed step in the same chunk.
To exploit the persistence of low-frequency structure during steady
interaction, we further introduce a frequency-mixed history prior guidance
that incorporates complementary information from the preceding clean latent
without an additional DiT forward pass.
Evaluated on Matrix-Game~2.0 and LingBot-World~v1/v2, CtrlCache achieves
1.21$\times$ to 1.41$\times$ DiT-backbone speedups without model retraining
while improving WBench Overall
scores over original inference across all three models.
\end{abstract}

\begin{figure}[!ht]
  \centering
  \includegraphics[width=0.68\textwidth]{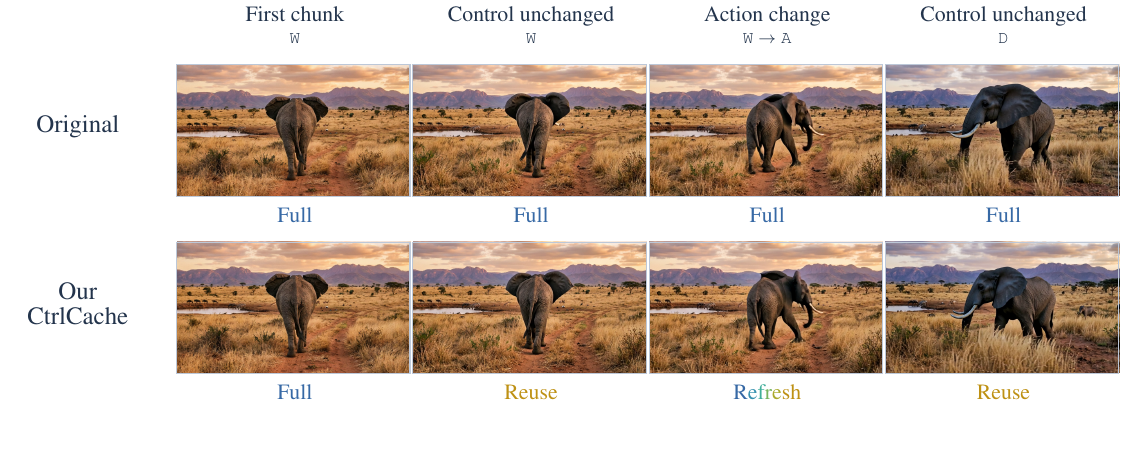}%
  \hfill
  \includegraphics[width=0.30\textwidth]{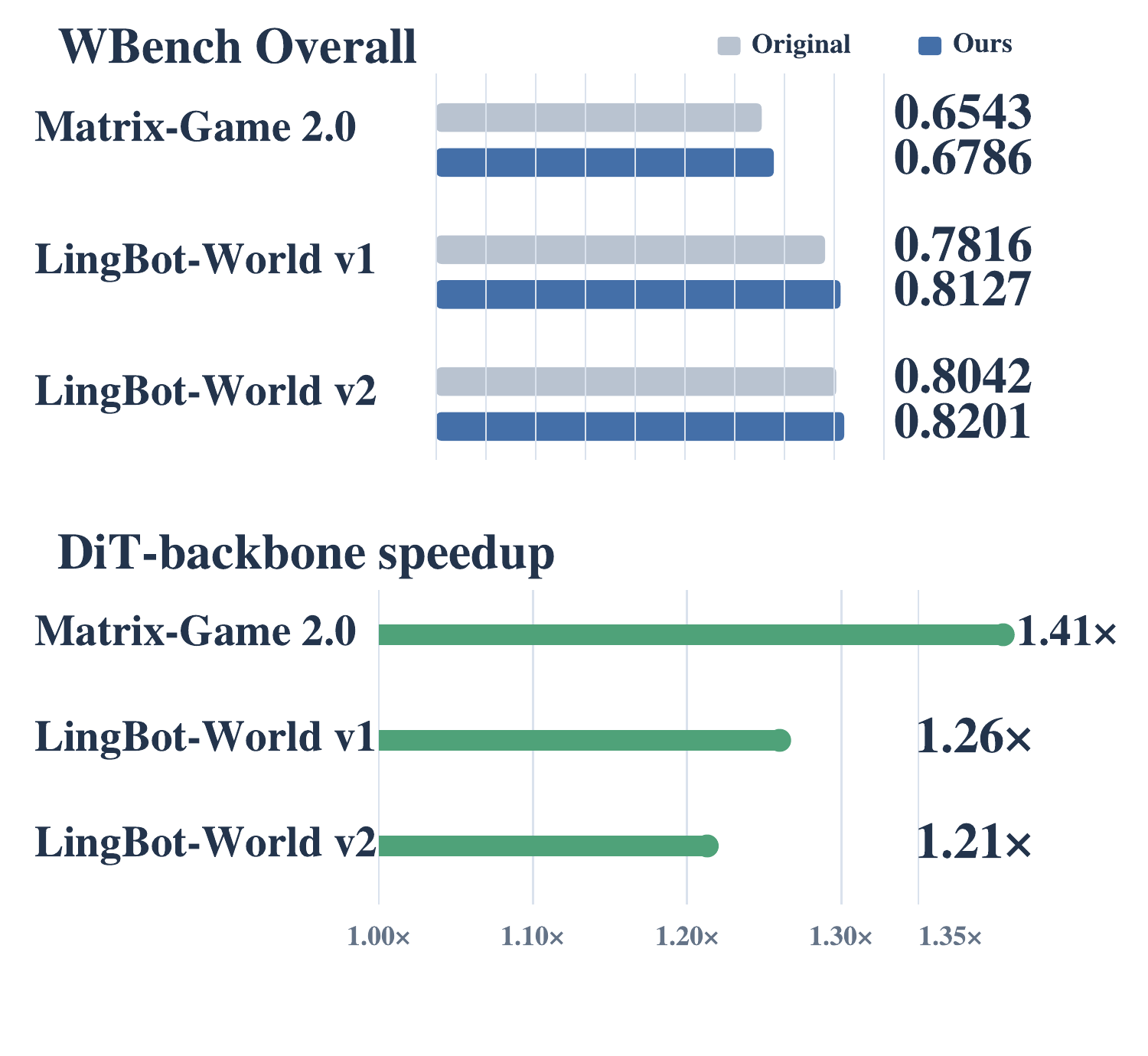}
  \vspace{-10pt}
  \caption{
  Our CtrlCache adapts DiT computation to the incoming controls. The left panel shows how the control-aware policy selects
  full computation, residual reuse, or refresh across representative control
  states. The right panel summarizes WBench Overall and DiT-backbone speedup on
  Matrix-Game~2.0~\citep{he2025matrix}, LingBot-World~v1~\citep{robbyant2026lingbot}, and LingBot-World~v2~\citep{gao2026lingbotv2}. Here, \emph{Full}
  denotes a regular full DiT forward, \emph{Reuse} denotes residual reuse at
  the selected denoising step, and \emph{Refresh} denotes a full forward at an
  action transition.
  }
  \label{fig:teaser}
\end{figure}

\section{Introduction}
\label{sec:introduction}

\begin{figure}[t!]
  \centering
  \includegraphics[width=\textwidth]{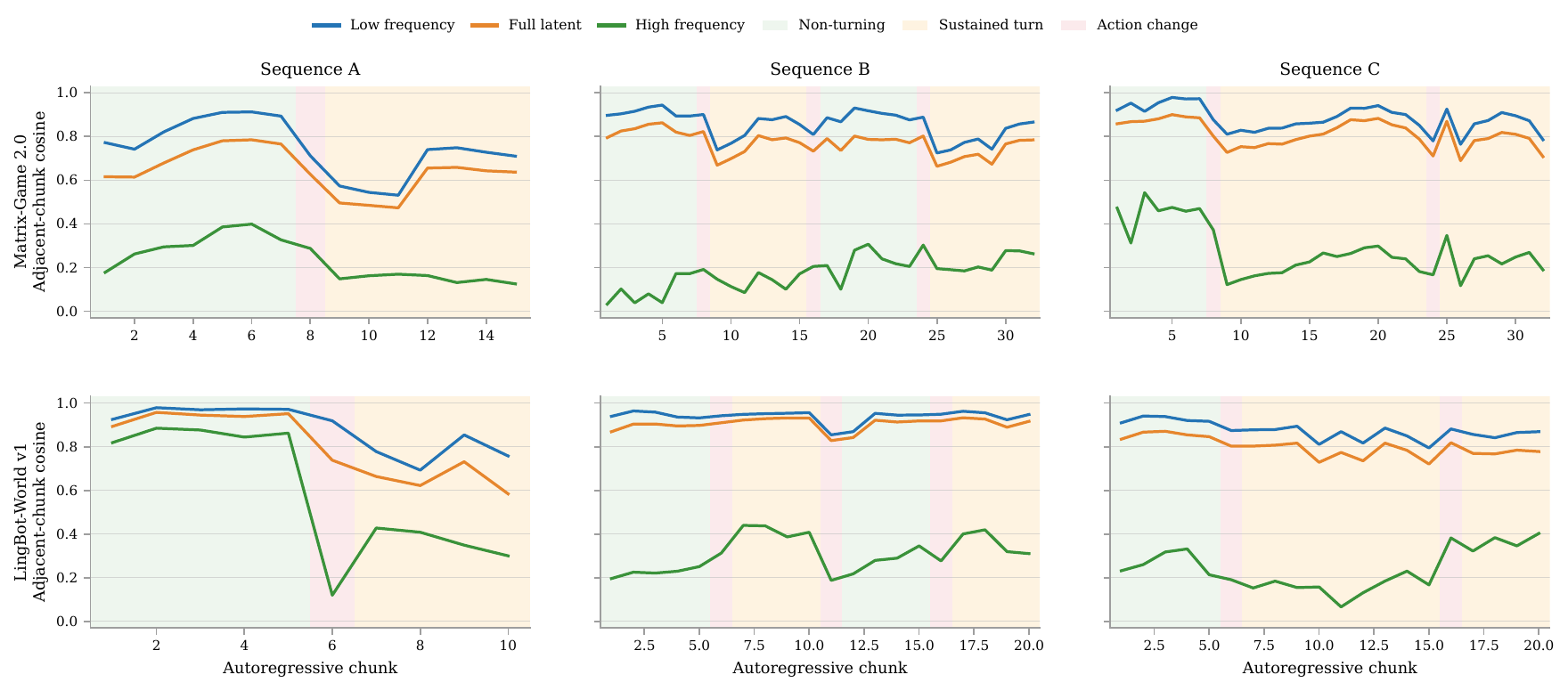}
  \vspace{-8pt}
  \caption{
  Cosine similarity between the clean latents of adjacent chunks
  under different control
  signals for three representative sequences.
  Each panel compares the full latent, low-frequency and high-frequency components.
  The shading background marks the control regime of each chunk, \emph{i.e.,} non-turning (steady), sustained turning,
  and action change (transition).
  Similarity drops at the action changes, and
  the low-frequency component stays more persistent than the high-frequency one.
  }
  \label{fig:control_state_layout}
  \vspace{-7pt}
\end{figure}

Interactive video world models turn a stream of user controls into continuously evolving
visual environments, offering a foundation for interactive simulation and
virtual exploration. Recent methods~\citep{valevski2025gamengen,feng2025thematrix,he2025matrix,robbyant2026lingbot,gao2026lingbotv2,sun2025worldplay} have made substantial progress toward
realistic, real-time, and long-horizon
interaction. However, efficient
inference remains essential since delays in generating each new segment directly
affect how promptly users can interact with the simulated world.

Many recent interactive world models~\citep{yin2025causvid,huang2025selfforcing,he2025matrix,robbyant2026lingbot} generate video autoregressively in short chunks and adopt few-step diffusion to reduce sampling cost.
Yet each chunk still requires multiple expensive full forward through
the large diffusion transformer (DiT) backbone.
Training-free caching
can reduce this repeated computation by reusing intermediate computation,
ranging from
fixed reuse schedules~\citep{ma2024deepcache,selvaraju2024fora,zhao2025pab} to adaptive ones driven by internal feature changes~\citep{liu2025teacache,zhou2025easycache}.
However, these methods were primarily developed for image or video
generation, and their reuse policies do not explicitly account for
action transitions during interactive generation.

This motivates examining how the incoming control sequence can guide
computation allocation and the use of generated history.
We first analyze the similarity between the clean latent representations
of adjacent chunks under different control regimes, and
make two observations as shown in Figure~\ref{fig:control_state_layout}.
First, latent similarity drops around action changes, suggesting
that
reuse policies should be applied more conservatively during these transitions.
Second, even during steady interaction, low-frequency components are
more persistent across chunks than high-frequency components,
suggesting that coarse scene layout carries over more reliably than
fine-grained appearance details.

These observations naturally raise two complementary design questions:
\emph{when} denoising cache should be reused or refreshed,
and \emph{what} content from the cached history is safer to carry forward if reusing them.
To this end, we introduce \textbf{CtrlCache}, a
training-free framework that couples control-aware computation allocation
with selective history cache transfer.
To decide \emph{when}, we introduce an \emph{action-aware scheduling and refresh policy}.
It first assigns each chunk to one of four control states, $\operatorname{initial}$, $\operatorname{transition}$, $\operatorname{turning}$, or $\operatorname{steady}$, based on the incoming control sequence.
Then, at one interior denoising step, initial chunks execute the full DiT computation to initialize the cache, transition chunks do so to refresh it, and
turning and steady chunks instead permit reuse using the transformer residual cache from the
most recent fully denoising step in
the same chunk.
In this way, the denoising cache can be selectively reused according to the incoming controls.
This policy is applied both across chunk boundaries and within each chunk, since a single chunk spans
multiple action samples.
Transferring the preceding chunk's clean latent frame without distinction, however, can
carry over fine details that no longer fit the evolving scene. Guided by
our frequency analysis, we further introduce a \emph{frequency-mixed history
prior guidance} to address \emph{what}.
The guidance builds a history prior from that frame, which emphasizes low-frequency structure
of that frame,
attenuates its high-frequency detail, and
gradually reduces the history prior contribution at later positions in the new
chunk.
The prior is then blended into the clean latent estimate after the first denoising step
to stabilize the structural layout of subsequent chunks under acceleration,
without an additional computational cost.
We enable this guidance only for
steady chunks, since during control transitions and sustained turning,
the geometry inherited from the preceding chunk may become outdated as
the viewpoint changes.
Together, these designs adapt both computation reuse and history
transfer to the current interaction state without retraining the model.

We evaluate CtrlCache on Matrix-Game~2.0~\citep{he2025matrix} and
LingBot-World~v1/v2~\citep{robbyant2026lingbot,gao2026lingbotv2}
using the WBench navigation track~\citep{ying2026wbench}.
Across all three backbones, CtrlCache achieves the highest WBench
Overall among all evaluated methods while delivering DiT-backbone
speedups ranging from $1.21\times$ to $1.41\times$.
Among the accelerated methods, it also achieves the highest PSNR and
SSIM and the lowest LPIPS on every backbone, demonstrating the
strongest fidelity to the Original outputs. Overall, our main contributions are:
\begin{itemize}[leftmargin=*,itemsep=0pt]
\vspace{-2mm}
    \item We analyze clean-latent similarity between adjacent chunks across
    control regimes, showing that similarity drops around action changes,
    while low-frequency components remain more persistent than high-frequency
    components during steady interaction.
    \item We propose CtrlCache, a training-free control-aware caching framework
    that uses action-aware scheduling and refresh to determine \emph{when} to
    reuse denoising computation, and a frequency-mixed history prior to select
    \emph{what} historical cache information to transfer during steady interaction.
    \item Extensive experiments on three interactive video world models demonstrate that CtrlCache accelerates the inference computation while improving WBench Overall scores, outperforming the evaluated caching baselines at comparable latency.
\end{itemize}

\FloatBarrier

\section{Related Work}
\label{sec:related_work}

\myparagraph{Interactive Video World Models.} Interactive video world models
generate visual observations conditioned on controls throughout an ongoing
rollout. Early autoregressive systems model action-conditioned latent dynamics,
as in Genie~\citep{bruce2024genie} and iVideoGPT~\citep{wu2024ivideogpt};
diffusion world models extend this setting to action-conditioned games and
navigation~\citep{alonso2024diamond,valevski2025gamengen,bar2025navigation}.
Recent systems broaden control and scene coverage through multimodal or
open-world generation~\citep{che2025gamegenx,yu2025gamefactory,guo2025mineworld},
while CausVid~\citep{yin2025causvid} and Self Forcing
\citep{huang2025selfforcing} establish causal few-step diffusion for streaming
generation. Long-horizon interaction and memory are further explored by The
Matrix~\citep{feng2025thematrix}, Yume~\citep{mao2025yume},
Matrix-Game~2.0~\citep{he2025matrix}, LingBot-World
\citep{robbyant2026lingbot,gao2026lingbotv2}, WorldPlay
\citep{sun2025worldplay}, and Matrix-Game~3.0
\citep{wang2026matrixgame3}. Despite few-step denoising, each newly generated
chunk still requires multiple forward passes through the DiT backbone, creating
a need for acceleration that preserves both control responsiveness and
cross-chunk temporal continuity.

\begin{figure*}[t]
  \centering
  \includegraphics[
    width=0.98\textwidth,
  ]{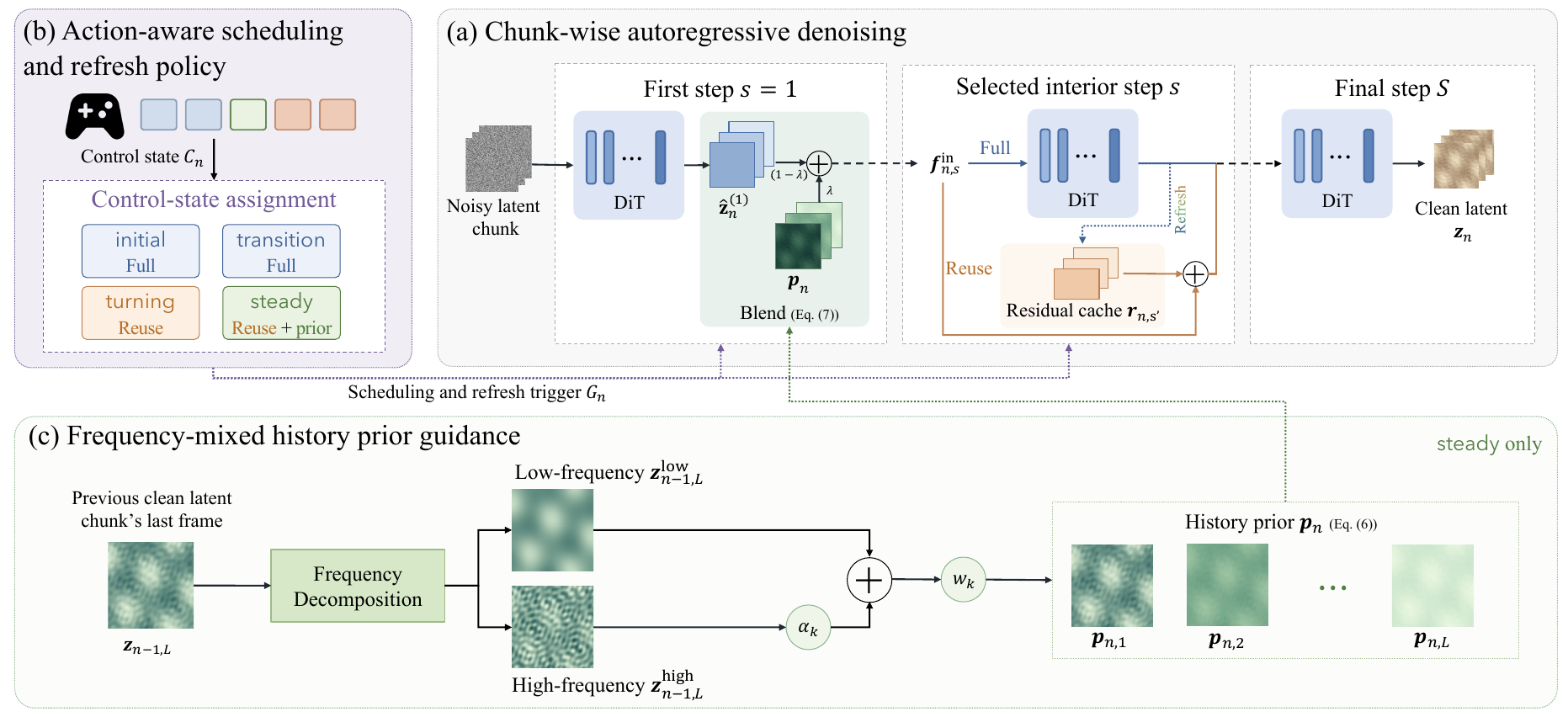}
  \caption{\textbf{Overview of CtrlCache.}
(a) Chunk-wise autoregressive denoising retains full computation at the
first and final steps.
(b) The action-aware policy enables residual reuse at the selected interior step
for turning and steady chunks, while preserving full computation for
initial and transition chunks.
(c) Frequency-mixed history prior guidance transfers structure from the
preceding chunk's last clean latent frame to guide the first-step
clean latent estimate during steady interaction.}
  \label{fig:method_overview}
  \vspace{-5pt}
\end{figure*}

\myparagraph{Training-Free Diffusion Inference Acceleration.}
Diffusion models generate images and videos through iterative denoising
\citep{ho2020ddpm,ho2022videodiffusion}, with each step requiring a backbone
forward pass; DiT~\citep{peebles2023dit} and related architectures therefore
offer substantial opportunities for intermediate-feature reuse. Training-free
caching methods use either predefined or adaptive schedules. Fixed schemes
include DeepCache~\citep{ma2024deepcache}, FORA~\citep{selvaraju2024fora},
$\Delta$-DiT~\citep{chen2024deltadit}, and PAB~\citep{zhao2025pab}, which reuse
features at selected network modules or timestep intervals. Adaptive schemes
include TeaCache~\citep{liu2025teacache}, EasyCache~\citep{zhou2025easycache},
FasterCache~\citep{lyu2025fastercache}, ToCa~\citep{zou2025toca}, and AdaCache
\citep{kahatapitiya2025adacache}, which adjust reuse according to feature
variation, token redundancy, or motion content. These methods reduce inference
cost without retraining. Recent work further studies structure-preserving
video caching and adaptive caching for video generation~\citep{fan2025taocache,agrawal2026acid}.
However, these works generally
ignore online action changes as
explicit cache-refresh signal.

FlowCache \citep{ma2026flowcache} extends caching to autoregressive video through independent chunk-wise
reuse decisions and bounded-history KV-cache compression, but does not interpret online action transitions.
X-Cache targets cross-chunk caching for autonomous-driving world models~\citep{zeng2026x},
a distinct application scenario from ours.
Light Interaction~\citep{lu2026lightinteraction} addresses interactive video
generation, yet ties denoising reuse to sufficiently similar historical views,
making it less suited to open-ended interactions with few or no revisits. Our method instead derives its reuse and refresh decisions directly from the control sequence.
Action transitions trigger full computation and
cache refresh, while other control states permit residual reuse without trajectory revisitation.

\section{CtrlCache}
\label{sec:method}

As illustrated in Figure~\ref{fig:method_overview}, CtrlCache builds on
established residual reuse (Sec.~\ref{sec:preliminaries}) and comprises two
complementary components. The \emph{action-aware scheduling and refresh policy}
(Sec.~\ref{sec:action_refresh}) uses the incoming controls to determine when to
reuse cached transformer residuals and when to restore full computation.
The \emph{frequency-mixed history prior guidance} (Sec.~\ref{sec:history_prior})
first constructs a history prior from the preceding chunk's last clean latent
frame, then uses this prior to guide the current chunk's first-step clean latent
estimate during steady interaction. Together, these components
determine when to reuse computation and what historical information to carry
forward.

\subsection{Preliminary}
\label{sec:preliminaries}
\myparagraph{Chunk-wise autoregressive denoising.}
Let $n$ index the $n$-th video chunk and $s\in\{1,\ldots,S\}$ as denoising steps within each chunk, where $S$ is the number of denoising steps per chunk.
To generate chunk $n$, the interactive video world models~\citep{he2025matrix,robbyant2026lingbot,gao2026lingbotv2} progressively denoise a noisy latent
conditioned on the incoming controls and previously generated history.
Each step performs a full forward pass through the
diffusion transformer (DiT), followed by a sampler update using the model
prediction. After the $S$ steps, the completed clean latent chunk provides
history for subsequent autoregressive generation.

\myparagraph{Residual reuse.}
\label{sec:residual_reuse}
Rather than computing the whole DiT transformer blocks at every denoising step, following \citep{liu2025teacache},
we adopt the established approximation of applying a cached residual from the most recent full forward to
the current step's input as the execution
mechanism underlying CtrlCache.
Let $\pmb{f}^{\mathrm{in}}_{n,s}$ and
$\pmb{f}^{\mathrm{out}}_{n,s}$ denote the feature tensors immediately before and after
the DiT transformer blocks for chunk $n$ at step $s$. A full forward process produces
the residual  $\pmb{r}_{n,s}=\pmb{f}^{\mathrm{out}}_{n,s}-\pmb{f}^{\mathrm{in}}_{n,s}$, which is cached for reuse within the current chunk.
At denoising step $s$, if the scheduling policy permits reuse and a residual
from an earlier fully computed step is cached, the current transformer-block
output $\widetilde{\pmb{f}}^{\mathrm{out}}_{n,s}$ can be approximated as:
\begin{equation}
  \widetilde{\pmb{f}}^{\mathrm{out}}_{n,s}
  =\pmb{f}^{\mathrm{in}}_{n,s}+\pmb{r}_{n,s'},
  \label{eq:residual_reuse}
\end{equation}
where $s'<s$ denotes the most recent preceding step $s'$ in the same chunk at which
the DiT was fully computed to obtain $\pmb{r}_{n,s'}$. The residual cache is refreshed
only after a full computation and remains unchanged during reuse,
allowing consecutive reuse steps to share the same residual.
The current input projection,
timestep embedding, and output head are still computed, while the cached residual
replaces the time-consuming transformer-block computation.

\subsection{Action-Aware Scheduling and Refresh Policy}
\label{sec:action_refresh}

As analyses in Figure~\ref{fig:control_state_layout}, the variation in adjacent chunk's clean-latent similarity across control regimes
motivates allocating computation according to the incoming controls. We thus propose an
action-aware scheduling and refresh policy to assign each chunk a control
state, to decide whether using residual reuse or full
computation at the selected interior denoising step.

\myparagraph{Control-state assignment.}
Based on the chunk's position in the rollout and the incoming controls, we
distinguish four control states, including $\operatorname{initial}$, $\operatorname{transition}$, $\operatorname{turning}$, and $\operatorname{steady}$.
The control state $C_n$ of the chunk $n$ is thus assigned as:
\begin{equation}
  C_n=
  \begin{cases}
    \operatorname{initial}, & n=1,\\
    \operatorname{transition}, & n>1,\ T_n=1,\\
    \operatorname{turning}, & n>1,\ T_n=0,\ R_n=1,\\
    \operatorname{steady}, & n>1,\ T_n=0,\ R_n=0.
  \end{cases}
  \label{eq:control_state}
\end{equation}
Here, $T_n$ indicates whether a control transition occurs, and $R_n$ indicates
whether a turning control is active. The first chunk is assigned the initial
state without requiring a preceding control input.
For subsequent chunks,
transition takes priority over turning, so a chunk containing an action
change is classified as transition even when a turning control is also active.
Steady state means unchanged, non-turning controls, while the generated scene may continue to evolve without remaining static.

Given these state definitions, the next step is to determine the indicators from the incoming
control sequence.
For a given backbone, let $\mathbf{a}_{n,j}$ denote the $j$-th
control input in the chunk $n$, where $j\in\{1,\ldots,J\}$ and $J$ is the number of
control inputs per chunk. We denote this sequence action by $\mathbf{a}_{n,1:J}$.
For $n>1$, we additionally define
$\mathbf{a}_{n,0}=\mathbf{a}_{n-1,J}$ as the last control input of the
preceding chunk, allowing the same  rule to detect changes both
across chunk boundaries and within a chunk. This auxiliary boundary action input is not included among the
current chunk's $J$ control inputs.

We distinguish a comparison between consecutive control inputs from a test of
the current turning-control magnitude. Let
$\operatorname{switch}(\mathbf{a},\mathbf{b})\in\{0,1\}$ indicate whether the
command changes between two inputs, and let
$\operatorname{rot}(\mathbf{a})\geq0$ denote the magnitude of the turning
control in an input.
For $n>1$, the transition indicators $T_n$ and the turning indicators $R_n$ are computed as:
\begin{equation}
  \begin{aligned}
    T_n
    &=\max_{1\leq j\leq J}
      \operatorname{switch}\!\left(
        \mathbf{a}_{n,j},\mathbf{a}_{n,j-1}
      \right),\\
    R_n
    &=\mathbbm{1}\!\left[
      \max_{1\leq j\leq J}\operatorname{rot}\!\left(\mathbf{a}_{n,j}\right)
      >\tau_{\mathrm{turn}}
    \right],
  \end{aligned}
  \label{eq:action_predicates}
\end{equation}
where $\mathbbm{1}[\cdot]$ denotes the indicator function, which returns
one when its argument is true and zero otherwise, and
$\tau_{\mathrm{turn}}$ is the backbone-specific turning threshold.
Checking $j=1$ detects a change at the chunk boundary, while checking
$j=\{2,\ldots,J\}$ captures changes within the chunk.

\myparagraph{Scheduling and refresh trigger.}
Given the control state $C_n$, we determine whether to enable residual reuse
at one selected interior denoising step $s\in\{2,\ldots,S-1\}$.
All remaining steps are fully computed, with the first initializing the
residual cache and the final preserving the model's refinement.
Let $G_n\in\{0,1\}$ indicate whether full computation is required at the
selected step:
\begin{equation}
  G_n=
  \begin{cases}
    1,   & n=1,\\
    T_n, & n>1.
  \end{cases}
  \label{eq:computation_gate}
\end{equation}
Here, $G_n=1$ triggers a cache refresh in which the DiT transformer blocks are fully
computed, and the resulting residual updates the current chunk's cache as the new residual cache.
When $G_n=0$, the transformer-block output at the selected step is
approximated using the cached residual via Eq.~(\ref{eq:residual_reuse}).
Initial chunks retain full computation to establish the rollout, and
transition chunks restore it at the selected step to accommodate changed
controls. Turning and steady chunks instead reuse the cached residual at
that step.

\subsection{Frequency-Mixed History Prior Guidance}
\label{sec:history_prior}
The action-aware scheduling and refresh policy controls residual reuse within each chunk. We further
introduce a frequency-mixed history prior guidance to 
determine what cached information is safer to carry forward during reusing. 
Here, we use information from the preceding chunk's
clean latent.
As the analyses in Figure~\ref{fig:control_state_layout}, even in the steady interaction regime,
low-frequency components encoding coarse scene layout remain more persistent
across chunks than high-frequency components encoding fine appearance details.
Transferring the preceding clean
latent without distinction can therefore impose details that no longer fit
the evolving scene.
To this end, we construct a history prior that
emphasizes coarse structure, and blend it into the current chunk's first-step clean
estimate. This guidance is enabled only when $C_n=\operatorname{steady}$ since
during transitions and sustained turning, the inherited geometry may itself
be changing with the viewpoint.

\myparagraph{Frequency-mixed history prior construction.}
Let $\pmb{z}_{n-1}$ denote the preceding chunk's clean latent, consisting of $L$ temporal latent frames. We use its
last frame, $\pmb{z}_{n-1,L}$, to construct the history prior
for the current chunk.
We decompose this latent into a low-frequency component
obtained by spatial average pooling and a high-frequency component defined
as the remaining difference:
\begin{equation}
  \begin{aligned}
    \pmb{z}^{\mathrm{low}}_{n-1,L}
    &=\operatorname{AvgPool}_{3\times3}(\pmb{z}_{n-1,L}),\\
    \pmb{z}^{\mathrm{high}}_{n-1,L}
    &=\pmb{z}_{n-1,L}-\pmb{z}^{\mathrm{low}}_{n-1,L}.
  \end{aligned}
  \label{eq:frequency_decomposition}
\end{equation}
Here, the $\mathrm{low}$ and $\mathrm{high}$ denote frequency
components. $\operatorname{AvgPool}_{3\times3}$ denotes spatial average pooling
with a $3\times3$ kernel, unit stride, and replicate padding, chosen for its
low computational cost. It operates independently on each latent frame and
channel, preserving spatial resolution without mixing information across the
temporal dimension.

The current chunk likewise contains $L$ temporal latent frames.
Position $k=1$ is closest in time to the
preceding chunk, and larger $k$ denotes a later position. Thus, $k$ indexes
positions within a latent chunk, while $s$ indexes denoising steps and $j$
indexes control inputs as shown in the previous sections. We construct a history prior latent frame $\pmb{p}_{n,k}$
at each position, with the same shape as $\pmb{z}_{n-1,L}$:
\begin{equation}
  \pmb{p}_{n,k}
  =w_k\bigl(
    \pmb{z}^{\mathrm{low}}_{n-1,L}
    +\alpha_k\pmb{z}^{\mathrm{high}}_{n-1,L}
  \bigr),
  \quad k=1,\ldots,L,
  \label{eq:history_prior}
\end{equation}
where $\alpha_k$ weights the high-frequency component, with $\alpha_1=1$
and $\alpha_k=\alpha\in[0,1]$ for $k\geq2$. The temporal weights $w_k$ satisfy
$1=w_1\geq w_2\geq\cdots\geq w_L\geq0$. With $\alpha_1=w_1=1$,
the first prior frame preserves the preceding chunk's last clean latent
frame in full. At later positions, $\alpha$ controls high-frequency
attenuation, while $w_k$ scales both frequency components according to
temporal distance. Stacking
$\{\pmb{p}_{n,1},\ldots,\pmb{p}_{n,L}\}$ along the temporal dimension yields the
history prior $\pmb{p}_n$ for chunk $n$. The model-specific values of
$\alpha$ and $w_k$ are reported in Section~\ref{sec:experimental_setup}.

\myparagraph{Guiding the clean latent estimate using the history prior.}
After constructing the history prior $\pmb{p}_n$ from the preceding
chunk, we use it to guide the current chunk's clean latent estimate.
Let $\widehat{\pmb{z}}_n^{(s)}$ denote the clean latent estimate of chunk $n$
obtained at denoising step $s$ using the base sampler's prediction
parameterization.
Both $\widehat{\pmb{z}}_n^{(s)}$ and $\pmb{p}_n$
span all $L$ temporal latent positions and have the same shape.
We then blend the
current estimate $\widehat{\pmb{z}}_n^{(s)}$ with the history prior $\pmb{p}_n$:
\begin{equation}
  \widehat{\pmb{z}}_n^{(s)}
  \leftarrow(1-\lambda)\widehat{\pmb{z}}_n^{(s)}+\lambda\pmb{p}_n,
  \label{eq:history_blend}
\end{equation}
where $\lambda\in[0,1]$ is the guidance strength and the arrow denotes an
in-place update of the clean estimate.
In this paper, we apply history prior guidance only at $s=1$, when the earliest
control-conditioned clean latent estimate of the current chunk becomes
available. This leaves the remaining denoising steps to adapt the transferred
structure to the current controls and refine visual details.
The sampler uses the updated clean estimate to construct the next noisy
latent under its original timestep and noise schedule. Thus, the guidance introduces no additional DiT forward computation cost.

\begin{table}[t!]
  \caption{
  Quality and efficiency comparison of CtrlCache and baselines on the WBench navigation track~\citep{ying2026wbench}. ``\emph{vs.} Original'' compares each accelerated method with the original
  full-computation model under identical prompts, controls, and initial frames. ``Efficiency'' reports DiT-backbone denoising latency
  (mean seconds per case), excluding KV-cache commit/update time, and speedup
  over Original. Bold denotes the best result within each model.
  }
  \label{tab:main_results}
  \centering
  \small
  \setlength{\tabcolsep}{2.0pt}
  \renewcommand{\arraystretch}{1.05}
  \resizebox{\textwidth}{!}{%
  \begin{tabular}{llcccccc|ccc|cc}
    \hline
    \multirow{2}{*}{Model} & \multirow{2}{*}{Methods} & \multicolumn{6}{c|}{WBench} & \multicolumn{3}{c|}{\emph{vs.} Original}
    & \multicolumn{2}{c}{Efficiency} \\
    \cline{3-8}\cline{9-11}\cline{12-13}
     &  & Quality$\uparrow$ & Cons.$\uparrow$ & Inter.$\uparrow$
    & Setting$\uparrow$ & Physics$\uparrow$ & Overall$\uparrow$
    & PSNR$\uparrow$ & SSIM$\uparrow$ & LPIPS$\downarrow$
    & Latency (s)$\downarrow$ & Speedup$\uparrow$ \\
    \hline
    \multirow{4}{*}{Matrix-Game 2.0} & Original  & 0.7360 & 0.6838 & 0.8261 & 0.5129 & 0.5127 & 0.6543
              & -- & -- & -- & 14.99 & 1.00$\times$ \\
    & TeaCache  & 0.7252 & 0.6820 & 0.8145 & 0.4982 & 0.5275 & 0.6495
              & 13.89 & 0.3344 & 0.4810 & 10.33 & 1.45$\times$ \\
    & EasyCache & 0.7245 & 0.6803 & 0.8093 & 0.5464 & 0.5518 & 0.6625
              & 13.46 & 0.3152 & 0.5021 & \textbf{10.17} & \textbf{1.47$\times$} \\
    & Our CtrlCache      & \textbf{0.7370} & \textbf{0.6867} & \textbf{0.8265} & \textbf{0.5527} & \textbf{0.5900} & \textbf{0.6786}
              & \textbf{14.51} & \textbf{0.3504} & \textbf{0.4757} & 10.67 & 1.41$\times$ \\
    \hline
    \multirow{4}{*}{LingBot-World v1} & Original  & 0.8073 & 0.8715 & 0.8088 & 0.7868 & 0.6338 & 0.7816
              & -- & -- & -- & 104.54 & 1.00$\times$ \\
    & TeaCache  & 0.8052 & \textbf{0.8755} & \textbf{0.8306} & 0.7844 & 0.7023 & 0.7996
              & 15.54 & 0.4638 & 0.3855 & 80.31 & 1.30$\times$ \\
    & EasyCache & \textbf{0.8074} & 0.8751 & 0.7907 & 0.7910 & 0.7287 & 0.7986
              & 15.26 & 0.4506 & 0.3946 & \textbf{79.80} & \textbf{1.31$\times$} \\
    & Our CtrlCache      & 0.8073 & 0.8737 & 0.8200 & 0.8246 & \textbf{0.7379} & \textbf{0.8127}
              & \textbf{17.27} & \textbf{0.5014} & \textbf{0.3451} & 83.03 & 1.26$\times$ \\
    \hline
    \multirow{4}{*}{LingBot-World v2} & Original  & 0.8290 & 0.8716 & \textbf{0.8389} & 0.7692 & 0.7124 & 0.8042
              & -- & -- & -- & 66.93 & 1.00$\times$ \\
    & TeaCache  & \textbf{0.8307} & \textbf{0.8848} & 0.8279 & 0.8078 & 0.7171 & 0.8137
              & 14.98 & 0.4176 & 0.3538 & 51.54 & 1.30$\times$ \\
    & EasyCache & 0.8303 & 0.8824 & 0.8296 & 0.8047 & 0.6983 & 0.8091
              & 14.90 & 0.4161 & 0.3586 & \textbf{51.12} & \textbf{1.31$\times$} \\
    & Our CtrlCache      & 0.8278 & 0.8789 & 0.8334 & \textbf{0.8306} & \textbf{0.7298} & \textbf{0.8201}
              & \textbf{18.35} & \textbf{0.5238} & \textbf{0.2832} & 55.18 & 1.21$\times$ \\
    \hline
  \end{tabular}}
\end{table}

\subsection{Overall mechanism}
For each chunk, CtrlCache first determines its control state from the
incoming controls and then coordinates computation reuse and history transfer
accordingly. Initial and transition chunks retain full computation throughout
the denoising process, whereas turning and steady chunks reuse a cached
residual at one selected interior step. The first and final denoising steps
remain fully computed in all cases. In addition, steady chunks apply the
frequency-mixed history prior to the clean latent estimate at $s=1$ to promote
structural consistency across chunks. After generating the clean latent chunk
$\pmb{z}_n$, CtrlCache retains its last temporal frame $\pmb{z}_{n,L}$ as the
history reference for the next chunk. The complete inference pseudocode is
provided in Appendix~\ref{sec:algo}.

\section{Experiments}
\label{sec:experiments}

\subsection{Experimental Setup}
\label{sec:experimental_setup}

\myparagraph{Models and baselines.}
We evaluate CtrlCache on three interactive video world models:
Matrix-Game~2.0~\citep{he2025matrix},
LingBot-World~v1~\citep{robbyant2026lingbot}, and
LingBot-World~v2~\citep{gao2026lingbotv2}, using their few-step,
chunk-wise autoregressive inference configurations. We compare against
the unmodified inference procedure (Original) and two representative
training-free caching methods,
TeaCache~\citep{liu2025teacache} and EasyCache~\citep{zhou2025easycache}.
Both baselines follow their own published reuse rules with own original denoising schedules, and neither uses our
frequency-mixed history prior guidance.

\myparagraph{Benchmark and evaluation metrics.}
We evaluate interactive generation on the navigation track of
WBench~\citep{ying2026wbench}, reporting video quality (Quality), consistency (Cons.),
interaction adherence (Inter.), setting adherence (Setting), and physics compliance (Physics).
The Overall score is the unweighted mean of these five dimension scores.
To assess fidelity to the unmodified model, we also report PSNR,
SSIM~\citep{wang2004image}, and LPIPS~\citep{zhang2018lpips} against the
paired Original. For efficiency, we report mean DiT-backbone
latency and speedup relative to Original.

\myparagraph{Implementation details.}
All experiments use a single NVIDIA A100 GPU with random seed $42$.
For each backbone, all methods share the same prompts, control sequences,
initial frames, and decoding settings, and retain the original denoising
The selected reuse step $s$ is $2$, $3$, $2$ for Matrix-Game~2.0,
LingBot-World~v1 and v2; reuse is enabled only
when permitted by $G_n$ (Eq.~(\ref{eq:computation_gate})).
For the turning indicator $R_n$ in Eq.~(\ref{eq:action_predicates}), we set $\tau_{\mathrm{turn}}=10^{-6}$ for
the mouse-control magnitude in Matrix-Game~2.0 and
$\tau_{\mathrm{turn}}=0$ for the yaw/pitch motion magnitude in both
LingBot-World variants.
Unless otherwise stated, history prior guidance uses $\alpha=0.5$ in Eq.~(\ref{eq:history_prior}) and
$\lambda=0.5$ in Eq.~(\ref{eq:history_blend}).
The temporal weights $(w_1,\ldots,w_L)$ are $(1,0.5,0.25)$ for
Matrix-Game~2.0 and LingBot-World~v1 ($L=3$), and $(1,0.5,0.25,0.125)$ for
LingBot-World~v2 ($L=4$).
More details are in Appendix~\ref{sec:supp_implementation}.

\begin{figure}[!t]
  \centering
  \begin{adjustbox}{max totalsize={1.0\textwidth}{0.98\textheight}}
    \begin{minipage}{\textwidth}
      \centering
      \setlength{\parskip}{-5pt}
      \includegraphics[width=\linewidth]{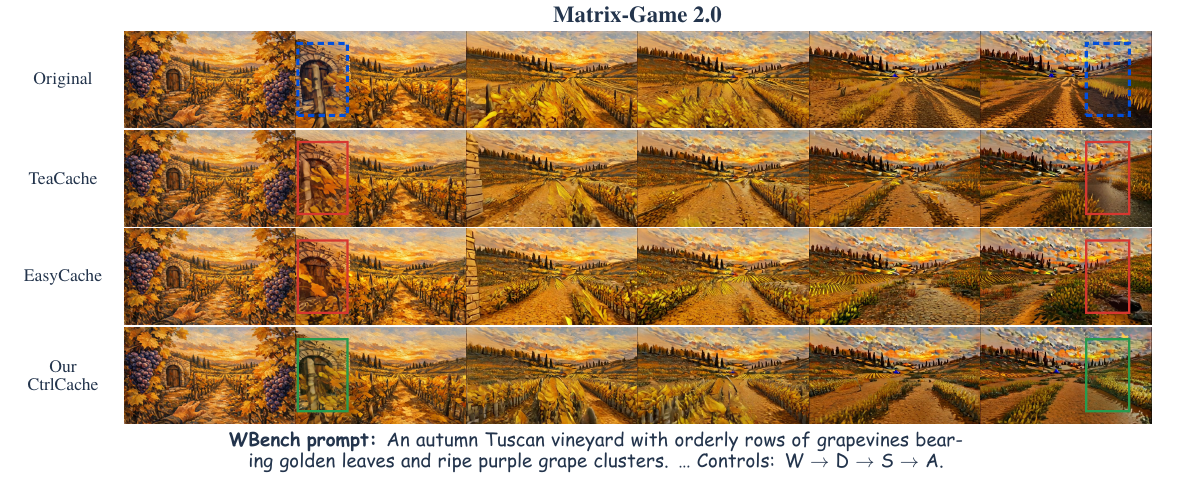}\par
      \vspace{-5pt}
      \includegraphics[width=\linewidth]{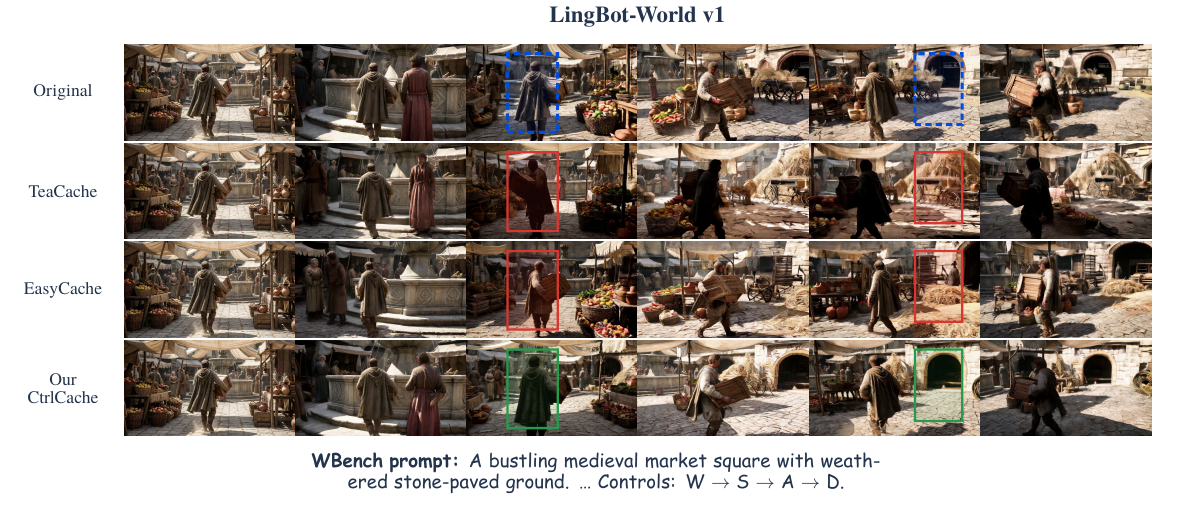}\par
      \vspace{-5pt}
      \includegraphics[width=\linewidth]{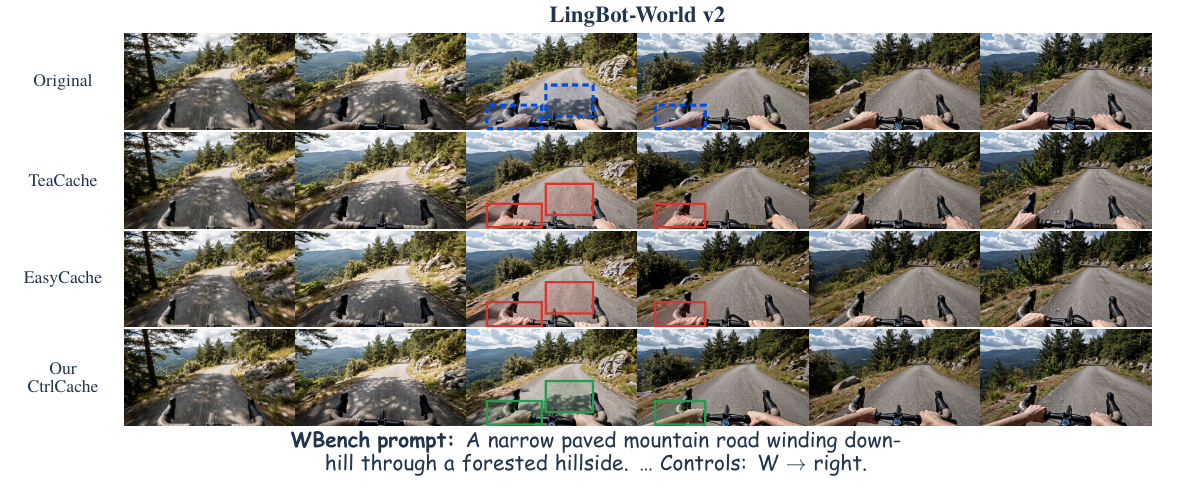}
    \end{minipage}
  \end{adjustbox}
  \vspace{-0.4cm}
  \caption{Qualitative comparisons of CtrlCache and baselines. For each
  case, all methods use the same initial condition and control sequence. The
  \textcolor{blue}{blue dashed rectangles} mark reference regions for comparing
  layout, object placement, and motion-related details across methods.
  The \textcolor{red}{red rectangles} highlight distortions in local appearance and
  scene structure, while the \textcolor{green!50!black}{green  rectangles}
  highlight better-preserved details in our results.
  Control strings follow the original WBench convention, where
  \texttt{W} / \texttt{S}/ \texttt{A} / \texttt{D}
  denote translational commands and \texttt{left} / \texttt{right} / \texttt{up} / \texttt{down} denote view
  changes.}
  \label{fig:qualitative_comparison}
\end{figure}

\begin{table}[!t]
  \caption{Ablation studies on Matrix-Game~2.0 and LingBot-World v1 using WBench navigation track. The first two variants evaluate naive residual reuse
(Eq.~(\ref{eq:residual_reuse})) alone and with our proposed action-aware
scheduling and refresh policy (abbreviated as \emph{Action-aware policy};
Sec.~\ref{sec:action_refresh}), respectively. The last
  three fix both components and apply the frequency-mixed history prior guidance (abbreviated as \emph{History prior};
Sec.~\ref{sec:history_prior}) on $\operatorname{steady}$ (St.), $\operatorname{steady}$
  and $\operatorname{turning}$ (St.+Tu.), or $\operatorname{steady}$, $\operatorname{turning}$ and $\operatorname{transition}$ (St.+Tu.+Tr.) chunks. Efficiency reports DiT-backbone denoising latency and speedup
  relative to Original. Bold denotes the best result within each model.}
  \label{tab:ablation}
  \centering
  \small
  \setlength{\tabcolsep}{3.5pt}
  \resizebox{\textwidth}{!}{%
  \begin{tabular}{llcccccc|cc}
    \hline
    \multirow{2}{*}{Model} & \multirow{2}{*}{Variant}
    & \multicolumn{6}{c|}{WBench} & \multicolumn{2}{c}{Efficiency} \\
    \cline{3-8}\cline{9-10}
    & & Quality$\uparrow$ & Cons.$\uparrow$ & Inter.$\uparrow$
    & Setting$\uparrow$ & Physics$\uparrow$ & Overall$\uparrow$
    & Latency (s)$\downarrow$ & Speedup$\uparrow$ \\
    \hline
    \multirow{6}{*}{Matrix-Game 2.0}
      & {Original} & 0.7360 & 0.6838 & 0.8261 & 0.5129 & 0.5127 & 0.6543 & 14.99 & 1.00$\times$ \\
      \cline{2-10}
      & + Residual reuse %
      & 0.7353 & \textbf{0.6870} & \textbf{0.8418} & 0.5285 & 0.4843 & 0.6554
      & 10.31 & 1.45$\times$ \\
      & + Action-aware policy
      & 0.7310 & 0.6802 & 0.8372 & 0.5417 & 0.5169 & 0.6614
      & 10.39 & 1.44$\times$ \\
      & + History prior (St.+Tu.+Tr.) & 0.7299 & 0.6772 & 0.8014 & 0.5436 & 0.5272 & 0.6559
      & 10.65 & 1.41$\times$ \\
      & + History prior (St.+Tu.) & 0.7301 & 0.6816 & 0.8200 & 0.5030 & 0.5856 & 0.6641
      & 10.67 & 1.41$\times$ \\
      & + History prior (St.) (Full CtrlCache) & \textbf{0.7370} & 0.6867 & 0.8265 & \textbf{0.5527}
      & \textbf{0.5900} & \textbf{0.6786} & 10.67 & 1.41$\times$ \\
    \hline
    \multirow{6}{*}{LingBot-World v1}
    & {Original} & 0.8073 & 0.8715 & 0.8088 & 0.7868 & 0.6338 & 0.7816 & 104.54 & 1.00$\times$ \\
      \cline{2-10}
      & + Residual reuse %
      & 0.8050 & \textbf{0.8756} & 0.8119 & 0.7815 & 0.7364 & 0.8021
      & 80.35 & 1.30$\times$ \\
      & + Action-aware policy
      & 0.8073 & 0.8688 & 0.8120 & 0.8148 & 0.7246 & 0.8055 & 82.98 & 1.26$\times$ \\
      & + History prior (St.+Tu.+Tr.) & \textbf{0.8076} & 0.8689 & 0.7845 & 0.7804 & 0.7040 & 0.7891
      & 83.05 & 1.26$\times$ \\
      & + History prior (St.+Tu.) & 0.8074 & 0.8716 & 0.7870 & 0.7893 & 0.7175 & 0.7946
      & 83.03 & 1.26$\times$ \\
      & + History prior (St.) (Full CtrlCache) & 0.8073 & 0.8737 & \textbf{0.8200} & \textbf{0.8246} & \textbf{0.7379}
      & \textbf{0.8127} & 83.03 & 1.26$\times$ \\
    \hline
  \end{tabular}}
  \vspace{-0.5cm}
\end{table}

\subsection{Main Results}
\label{sec:main_results}

Table~\ref{tab:main_results} reports the quantitative comparison on all three
backbones. Overall, CtrlCache attains the best WBench Overall score and the
closest agreement with the original model on every backbone, at a speedup
comparable to the caching baselines.
On Matrix-Game~2.0, CtrlCache is the only method that improves over Original
on all five WBench dimensions, raising Overall from 0.6543 to 0.6786 and
surpassing the relatively best EasyCache (0.6625) while delivering
1.05 higher PSNR.
On LingBot-World~v1/~v2, CtrlCache achieves best Overall, PSNR, SSIM and LPIPS, while keeping a comparable
1.26$\times$ and 1.21$\times$ speedup, respectively.
The caching baselines remain marginally faster than CtrlCache on every
backbone, since their reuse criteria are free to skip computation at action
transitions.
CtrlCache instead spends that computation on transition chunks,
which is what yields the highest setting adherence and physics compliance on
all three backbones.

Figure~\ref{fig:qualitative_comparison} presents qualitative comparisons with all three backbones.
For each case, all methods use identical
initial conditions and control sequences.
Although the overall scene
layouts remain similar, TeaCache and
EasyCache exhibit distortions in local appearance
and scene structure., \emph{e.g.,}
the disappearance of an arched doorway in LingBot-World~v1.
Our CtrlCache restores full computation at control transitions and better
preserves local details and scene structure. These qualitative results
are consistent with its higher PSNR and lower LPIPS reported in
Table~\ref{tab:main_results}.
Video comparisons are available on our project page: \url{\projecturl}.

\subsection{Ablation Study}
\label{sec:ablation_studies}

\myparagraph{Effect of naive residual reuse.}
We first evaluate naive residual reuse (Eq.~(\ref{eq:residual_reuse})) at the selected interior denoising
step $s$, without the action-aware policy or history prior guidance.
We reuse at $s=2,3$ for Matrix-Game~2.0, LingBot-World v1 respectively, based on the scheduling configuration of each model (see Appendix~\ref{sec:supp_implementation}), and leaves each model's final correction step fully evaluated.
This reduces transformer-block computation while slightly improving Overall on
both backbones. However, the gains are uneven
across dimensions, \emph{e.g.,} physics compliance on Matrix-Game~2.0 declining
relative to Original, suggesting that
naive residual reuse may compromise quality,
motivating a policy that explicitly accounts for control transitions.

\myparagraph{Effect of the action-aware scheduling and refresh policy.}
Building on naive residual reuse, we enable the action-aware policy in Sec.~\ref{sec:action_refresh} to
restore full computation at control transitions, while steady and turning states do residual reuse.
It improves Overall on both backbones, as well as setting and physics
compliance on Matrix-Game~2.0 and setting compliance on LingBot-World~v1.
Overall, the policy improves generation performance by better preserving scene
settings under acceleration.

\myparagraph{Effect of the frequency-mixed history prior guidance.}
We keep residual reuse and the action-aware policy fixed and vary the
control states in which history prior guidance is applied.
By default, we only apply the frequency-mixed history prior guidance (Sec.~\ref{sec:history_prior}) on the steady state.
Restricting
guidance to steady chunks yields the highest Overall scores among the
evaluated variants, while extending it to turning or transition chunks
reduces performance. These results suggest that
the prior is most beneficial during steady interaction, whereas carrying
forward historical structure during turning or control transitions may
interfere with the required scene changes.

\myparagraph{Hyperparameter analysis.}
We provide additional sensitivity analyses of the selected residual-reuse
step $s$ (Eq.~(\ref{eq:residual_reuse})), and history prior settings, including the temporal weights $w_k$ (Eq.~(\ref{eq:history_prior})), high-frequency component weight $\alpha$ (Eq.~(\ref{eq:history_prior})), and blend guidance
strength $\lambda$ (Eq.~(\ref{eq:history_blend})) in the Appendix~\ref{sec:supp_hyperparameter_study}.

\section{Conclusion}
\label{sec:conclusion}

We propose CtrlCache, a training-free framework that uses online controls
to guide both computation reuse and history transfer in interactive video
world models. Its action-aware scheduling and refresh policy controls
residual reuse and restores full computation at control transitions, while
frequency-mixed history prior guidance transfers persistent coarse structure
during steady interaction. Experiments on Matrix-Game~2.0 and
LingBot-World~v1/~v2 show that CtrlCache improves WBench Overall over original models
and
achieves
comparable speedups. These findings highlight the value of
adapting acceleration to the control regime and the persistence of
historical information.

\myparagraph{Limitation.} CtrlCache currently relies on backbone-specific action indicators
and a fixed reuse step that varies with each backbone's denoising schedule.
We will explore more general control representations and adaptive
step selection in the future work.

\bibliography{iclr2027_conference}
\bibliographystyle{iclr2027_conference}

\clearpage
\appendix

\counterwithin{figure}{section}
\counterwithin{table}{section}
\section*{Appendix}
This appendix provides CtrlCache's inference pseudocode (Sec.~\ref{sec:algo}), more implementation details (Sec.~\ref{sec:supp_implementation}),
evaluation protocol (Sec.~\ref{sec:supp_evaluation}), hyperparameter analysis (Sec.~\ref{sec:supp_hyperparameter_study}), and additional Qualitative Examples (Sec.~\ref{sec:supp_qualitative}).

\section{CtrlCache Algorithm}
\label{sec:algo}
Algorithm~\ref{alg:ctrlcache} presents the inference procedure of CtrlCache,
complementing the method description in Section~\ref{sec:method} of the
main paper. The pseudocode details how the action-aware scheduling and
refresh policy and frequency-mixed history prior guidance are integrated
into chunk-wise autoregressive generation.

\begingroup
\RestyleAlgo{ruled}
\LinesNumbered
\begin{algorithm}[h]
\footnotesize
\SetAlgoLined
\SetAlgoNlRelativeSize{-1}
\SetKwInOut{Input}{Input}
\SetKwInOut{Output}{Output}
\SetKwInput{Initialization}{Initialization}
\caption{CtrlCache Inference}
\label{alg:ctrlcache}
\Input{Pretrained world model and base sampler; initial conditioning and online
controls; $N$ chunks, each with $J$ control inputs, $L$ temporal latent frames,
and $S$ denoising steps; one selected reuse step in $\{2,\ldots,S-1\}$;
prior parameters $\alpha$, $\{w_k\}_{k=1}^{L}$, and $\lambda$.}
\Output{Clean latent chunks $\{\pmb{z}_n\}_{n=1}^{N}$.}
\Initialization{Initialize the base model's autoregressive context from the
initial conditioning. Set $\alpha_1=1$ and $\alpha_k=\alpha$ for $k\geq2$.}

\For{$n=1$ \KwTo $N$}{
  Read the current control inputs $\mathbf{a}_{n,1:J}$\;
  \If{$n>1$}{
    $\mathbf{a}_{n,0}\gets\mathbf{a}_{n-1,J}$\;
    Compute $T_n$ and $R_n$ via Eq.~(\ref{eq:action_predicates})\;
  }
  Assign $C_n$ and $G_n$ via Eqs.~(\ref{eq:control_state}) and
  (\ref{eq:computation_gate})\;
  Clear the within-chunk residual cache; initialize the noisy latent using
  the base sampler\;
  \If{$C_n=\operatorname{steady}$}{
    Decompose $\pmb{z}_{n-1,L}$ into low- and high-frequency components
    via Eq.~(\ref{eq:frequency_decomposition})\;
    Construct $\pmb{p}_{n,k}$ for $k=1,\ldots,L$ via
    Eq.~(\ref{eq:history_prior}); stack them along the temporal dimension
    to obtain $\pmb{p}_n$\;
  }
  \For{$s=1$ \KwTo $S$}{
    Prepare $\pmb{f}^{\mathrm{in}}_{n,s}$ and the current timestep and
    control conditioning\;
    \eIf{$G_n=0$ \textnormal{and} $s$ \textnormal{is the selected reuse step}}{
      Approximate $\widetilde{\pmb{f}}^{\mathrm{out}}_{n,s}$ using the
      latest cached residual $\pmb{r}_{n,s'}$ via
      Eq.~(\ref{eq:residual_reuse})\;
    }{
      Fully compute the transformer blocks to obtain
      $\pmb{f}^{\mathrm{out}}_{n,s}$\;
      Cache $\pmb{r}_{n,s}\gets\pmb{f}^{\mathrm{out}}_{n,s}
      -\pmb{f}^{\mathrm{in}}_{n,s}$; $s'\gets s$\;
    }
    Obtain $\widehat{\pmb{z}}_n^{(s)}$ from the full or approximated block
    output using the output head and the base prediction parameterization\;
    \If{$C_n=\operatorname{steady}$ \textnormal{and} $s=1$}{
      Blend the current estimate with the history prior: $\widehat{\pmb{z}}_n^{(1)}
      \gets(1-\lambda)\widehat{\pmb{z}}_n^{(1)}+\lambda\pmb{p}_n$\ via Eq.~(\ref{eq:history_blend});
    }
    Perform the base sampler update using $\widehat{\pmb{z}}_n^{(s)}$
    and the original timestep and noise schedule\;
  }
  Obtain the clean latent chunk $\pmb{z}_n$ from the final sampler output\;
  Retain $\pmb{z}_{n,L}$ as the history reference for the next chunk\;
  Update the base model's autoregressive context with $\pmb{z}_n$\;
}
\end{algorithm}
\endgroup

\section{Implementation Details}
\label{sec:supp_implementation}

\paragraph{Model-specific inference settings.}
Original inference uses denoising schedules $[1000,908,713]$ for Matrix-Game~2.0,
$[999,978,947,825]$ for LingBot-World v1, and $[999,967,908,768]$ for
LingBot-World v2. Our default residual-reuse timesteps are $908$, $947$, and
$967$, respectively. Unless otherwise noted, the history prior uses
frequency-mixing coefficient $\alpha=0.5$ and blend strength $\lambda=0.5$. Its temporal decay
is $[1,0.5,0.25]$ for Matrix-Game~2.0 and LingBot-World v1, and
$[1,0.5,0.25,0.125]$ for LingBot-World v2.

\textbf{Matrix-Game~2.0.}
For each control input $j$, the keyboard condition is a four-dimensional vector
$\mathbf{k}_j=$[\texttt{W,S,A,D}], where the four entries denote the activation values of
the \texttt{W,S,A,D} commands at input $j$.
The mouse condition is a two-dimensional vector
$\mathbf{m}_j=[m^{\mathrm{vertical}}_j,m^{\mathrm{horizontal}}_j]$; up/down
controls use the first entry and left/right controls use the second entry. In
the implementation, the corresponding mouse values are $\pm0.1$. The complete
action signature is $\mathbf{a}_j=(\mathbf{k}_j,\mathbf{m}_j)$. For the first
chunk we use the $\operatorname{initial}$ state because no previous chunk exists.
For every later chunk, a change between adjacent control inputs, including a
change across the preceding-chunk boundary or within the chunk, gives
$\operatorname{transition}$. If no such change occurs, the chunk is $\operatorname{turning}$ when
any control input satisfies
$\max_i|m_{j,i}|>10^{-6}$, where $i$ indexes the two mouse coordinates;
otherwise it is $\operatorname{steady}$. The state classifier uses the unramped mouse
command, so the smoothing applied before model inference is not mistaken for
an action transition.

\textbf{LingBot-World v1/v2.}
The two LingBot backbones share the same WBench navigation representation. At
control input $j$, the action signature is
$\mathbf{u}_j=(\mathbf{v}_j,y_j,p_j)$, where
$\mathbf{v}_j=[v_j^{\mathrm{fwd}},v_j^{\mathrm{right}}]$ encodes the
local-frame motion command, with components for forward/backward and right/left
motion; $y_j$ is yaw, and $p_j$ is pitch. The token mapping is
\texttt{W}$=[1,0]$, \texttt{S}$=[-1,0]$, \texttt{A}$=[0,-1]$, \texttt{D}$=[0,1]$ for $\mathbf{v}_j$;
\texttt{left}/\texttt{right} set $y_j=-1/+1$; and
\texttt{up}/\texttt{down} set $p_j=+1/-1$.
When no control transition occurs, a chunk is classified as
$\operatorname{turning}$ state if any control input has nonzero yaw or pitch; otherwise, it is
$\operatorname{steady}$ state.

For a concrete example, consider the WBench action sequence
$\texttt{W}\mathbin{\rightarrow}\texttt{left}\mathbin{\rightarrow}
\texttt{up}\mathbin{\rightarrow}\texttt{D}$, with each command held for one
action interval. The corresponding conditions are:
\[
\begin{array}{c|c|c}
\text{Action} &
\text{Matrix-Game~2.0 }(\mathbf{k}_j,\mathbf{m}_j) &
\text{LingBot-World v1/v2 }(\mathbf{v}_j,y_j,p_j) \\
\hline
\texttt{W} &
([1,0,0,0],[0,0]) &
([1,0],0,0) \\
\texttt{left} &
([0,0,0,0],[0,-0.1]) &
([0,0],-1,0) \\
\texttt{up} &
([0,0,0,0],[0.1,0]) &
([0,0],0,1) \\
\texttt{D} &
([0,0,0,1],[0,0]) &
([0,1],0,0)
\end{array}
\]

\section{Evaluation Protocol}
\label{sec:supp_evaluation}

\myparagraph{Evaluation split and paired runs.}
We sample 40 cases from the WBench navigation split and use generation seed
$42$. For each backbone, all methods are evaluated on the same sampled cases
with the same seed. Original and accelerated runs also share the prompts,
controls, initial frames, and decoding settings, so fidelity is computed on
paired outputs.

\myparagraph{WBench aggregation.}
We report five navigation-track dimensions that consist of Quality (the mean of six video
quality measures), Consistency (the mean of eight consistency measures),
Interaction (interaction adherence), Setting (the mean of scene and subject
adherence), and Physics (the mean of visual plausibility and causal fidelity).
WBench Overall is the unweighted arithmetic mean of these five dimension
scores. Each component is averaged over the cases for which that measure is
defined.

\myparagraph{Paired fidelity metrics.}
For each case, PSNR, SSIM, and LPIPS are computed frame by frame between the
accelerated video and its paired Original video. We first average each metric
over frames within a video, then report the unweighted mean of the resulting
per-case values. Thus, each case contributes equally to the reported fidelity
metrics.

\myparagraph{Efficiency evaluation.} For efficiency, we report mean DiT denoising latency and
speedup relative to Original. Timing uses CUDA synchronization and covers only DiT computation
during denoising, excluding KV-cache commit/update, model loading, condition preparation, VAE
decoding, and video encoding.

\section{Hyperparameter Study}
\label{sec:supp_hyperparameter_study}

We study residual-reuse timesteps and the history-prior hyperparameters
(frequency composition, blend, and temporal decay) on Matrix-Game~2.0 and
LingBot-World v1; results are summarized in Table~\ref{tab:hyperparameters}.
For the history-prior studies, we keep the default residual-reuse timestep and
action-aware scheduling fixed, and apply the prior only to steady chunks. The
reuse-timestep study isolates the selected residual-reuse step.

\myparagraph{Effect of the denoising step $s$ for residual reuse (Eq.~(\ref{eq:residual_reuse})).}
We evaluate residual reuse at each denoising step after the first, including the
final step. On Matrix-Game~2.0, reuse at the interior denoising step $s=2$, where
timestep is 908, achieves 0.6554 Overall, whereas reuse at the final denoising step $s=S=3$ (\emph{i.e.,} timestep 713) reduces
the score to 0.4930. On LingBot-World v1, the corresponding no-refresh
settings at $s=2,3,4$, \emph{i.e.,}
timesteps 978, 947, and 825, achieve 0.8014, 0.8021, and
0.5624, respectively. The higher scores at interior steps support reusing
an interior residual while fully executing the final denoising step in the
default configuration.

\myparagraph{Effect of the temporal weights $w_k$ (Eq.~(\ref{eq:history_prior})).}
With $\alpha=0.5$ and blend strength $\lambda=0.5$, the decayed temporal weights
$\{w_k\}=[1,0.5,0.25]$ outperform the uniform weights
$\{w_k\}=[1,1,1]$ on both Matrix-Game~2.0 and
LingBot-World v1, indicating that temporally decayed weighting is preferable
to uniform weighting.

\myparagraph{Effect of the frequency composition weight $\alpha_k$ and blend guidance strength $\lambda$ (Eq.~(\ref{eq:history_blend})).}
We vary one factor at a time on both Matrix-Game~2.0 and LingBot-World v1.
The best-performing configuration on both models uses
$\alpha=0.5$ and $\lambda=0.5$.
In the sweep, $\alpha_1=1$ is fixed, while $\alpha$ controls the
high-frequency weight for $k\geq2$.

\begin{table}[!ht]
  \caption{Hyperparameter studies on Matrix-Game~2.0 and LingBot-World v1.
  Each entry is WBench Overall; bold denotes the best result within each model
  and panel. (a)~Reuse denoising step $s$. (b)~Temporal weights $w_k$ ($\alpha=0.5$, $\lambda=0.5$).
  (c)~Frequency composition ($\lambda=0.5$). (d)~Blend strength $\lambda$ ($\alpha=0.5$).}
  \label{tab:hyperparameters}
  \centering
  \small
  \begin{minipage}[t]{0.48\textwidth}
    \centering
    {(a) Effect of the denoising step $s$ for residual reuse}\\[-0.25ex]
    \setlength{\tabcolsep}{2pt}
    \begin{tabular}{lccc}
      \hline
      Model & Step $s$ & Timestep & Overall$\uparrow$ \\
      \hline
      \multirow{2}{*}{Matrix-Game 2.0}
        & 2 & 908 & \textbf{0.6554} \\
        & 3 & 713 & 0.4930 \\
      \hline
      \multirow{3}{*}{LingBot-World v1}
        & 2 & 978 & 0.8014 \\
        & 3 & 947 & \textbf{0.8021} \\
        & 4 & 825 & 0.5624 \\
      \hline
    \end{tabular}
  \end{minipage}
  \hfill
  \begin{minipage}[t]{0.48\textwidth}
    \centering
    {(b) Effect of the temporal weights $w_k$}\\[-0.25ex]
    \setlength{\tabcolsep}{7pt}
    \begin{tabular}{lcc}
      \hline
      Model & $w_k$ & Overall$\uparrow$ \\
      \hline
      \multirow{2}{*}{Matrix-Game 2.0}
        & $[1,1,1]$ & 0.6594 \\
        & $[1,0.5,0.25]$ & \textbf{0.6786} \\
      \hline
      \multirow{2}{*}{LingBot-World v1}
        & $[1,1,1]$ & 0.8006 \\
        & $[1,0.5,0.25]$ & \textbf{0.8127} \\
      \hline
    \end{tabular}
  \end{minipage}

  \vspace{1.2ex}
  \begin{minipage}[t]{0.48\textwidth}
    \centering
    {(c) Effect of the frequency composition weight $\alpha$}\\[-0.25ex]
    \setlength{\tabcolsep}{7pt}
    \begin{tabular}{lccc}
      \hline
      Model & $\alpha$ & $\lambda$ & Overall$\uparrow$ \\
      \hline
      \multirow{3}{*}{Matrix-Game 2.0}
        & 0.0 & 0.5 & 0.6578 \\
        & 0.5 & 0.5 & \textbf{0.6786} \\
        & 1.0 & 0.5 & 0.6552 \\
      \hline
      \multirow{3}{*}{LingBot-World v1}
        & 0.0 & 0.5 & 0.7931 \\
        & 0.5 & 0.5 & \textbf{0.8127} \\
        & 1.0 & 0.5 & 0.7923 \\
      \hline
    \end{tabular}
  \end{minipage}
  \hfill
  \begin{minipage}[t]{0.48\textwidth}
    \centering
    {(d) Effect of the blend guidance strength $\lambda$}\\[-0.25ex]
    \setlength{\tabcolsep}{7pt}
    \begin{tabular}{lccc}
      \hline
      Model & $\alpha$ & $\lambda$ & Overall$\uparrow$ \\
      \hline
      \multirow{3}{*}{Matrix-Game 2.0}
        & 0.5 & 0.25 & 0.6604 \\
        & 0.5 & 0.50 & \textbf{0.6786} \\
        & 0.5 & 1.00 & 0.6724 \\
      \hline
      \multirow{3}{*}{LingBot-World v1}
        & 0.5 & 0.25 & 0.7964 \\
        & 0.5 & 0.50 & \textbf{0.8127} \\
        & 0.5 & 1.00 & 0.7909 \\
      \hline
    \end{tabular}
  \end{minipage}
\end{table}

\section{Additional Qualitative Examples}
\label{sec:supp_qualitative}

We provide additional qualitative comparisons for all three backbones. Each
comparison contains six successive frames generated from the same initial
condition and control sequence, with rows corresponding to Original, TeaCache,
EasyCache, and Our CtrlCache. Controls use the original WBench tokens,
[\texttt{W,S,A,D}], for translational commands and left/right/up/down for view changes.
The corresponding videos are available on our project page: \url{\projecturl}.

\clearpage
\begin{figure}[p]
  \centering
  \makebox[\textwidth][c]{%
    \includegraphics[width=1.1\textwidth]{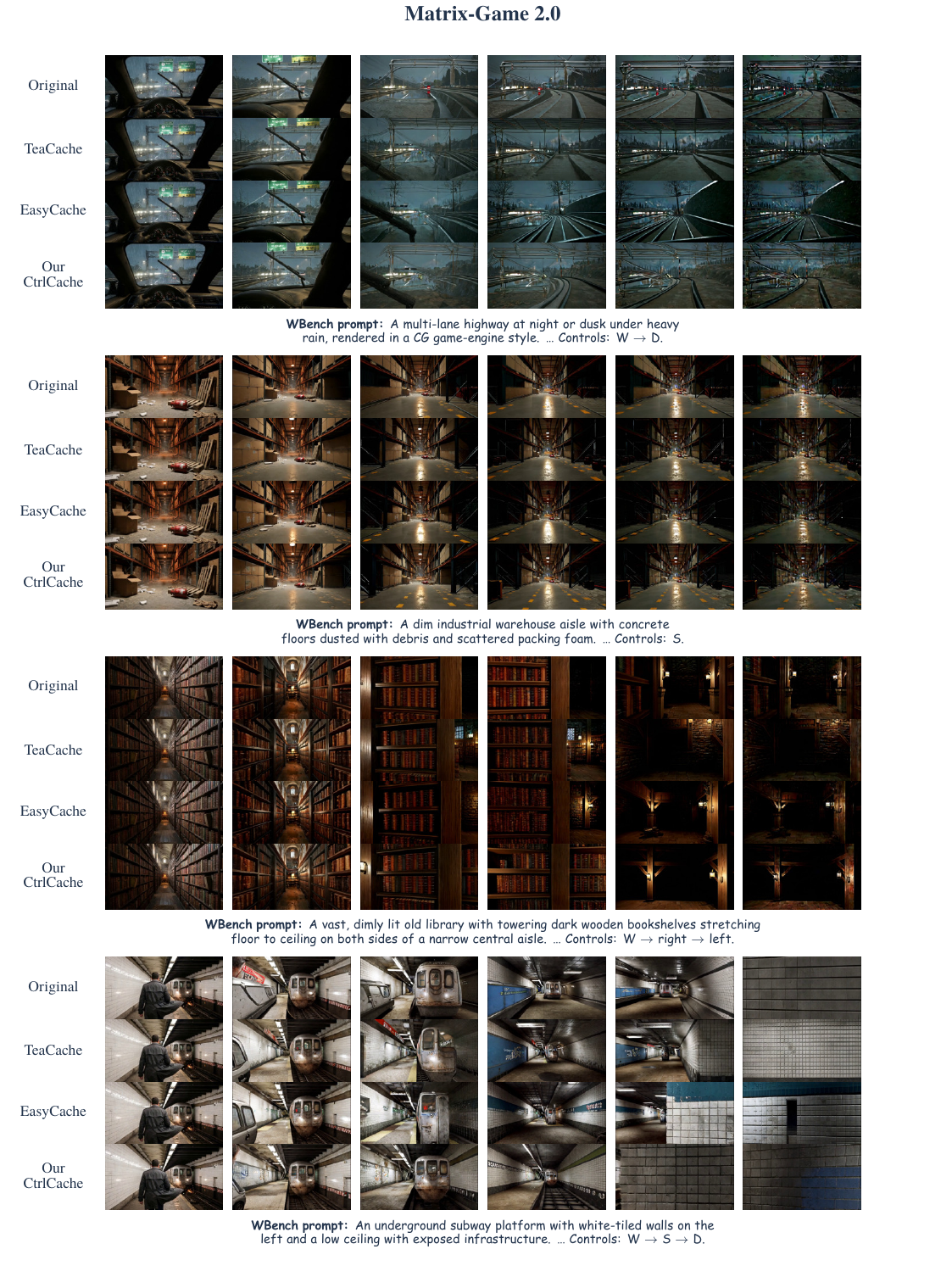}}
\end{figure}

\clearpage
\begin{figure}[p]
  \centering
  \makebox[\textwidth][c]{%
    \includegraphics[width=1.1\textwidth]{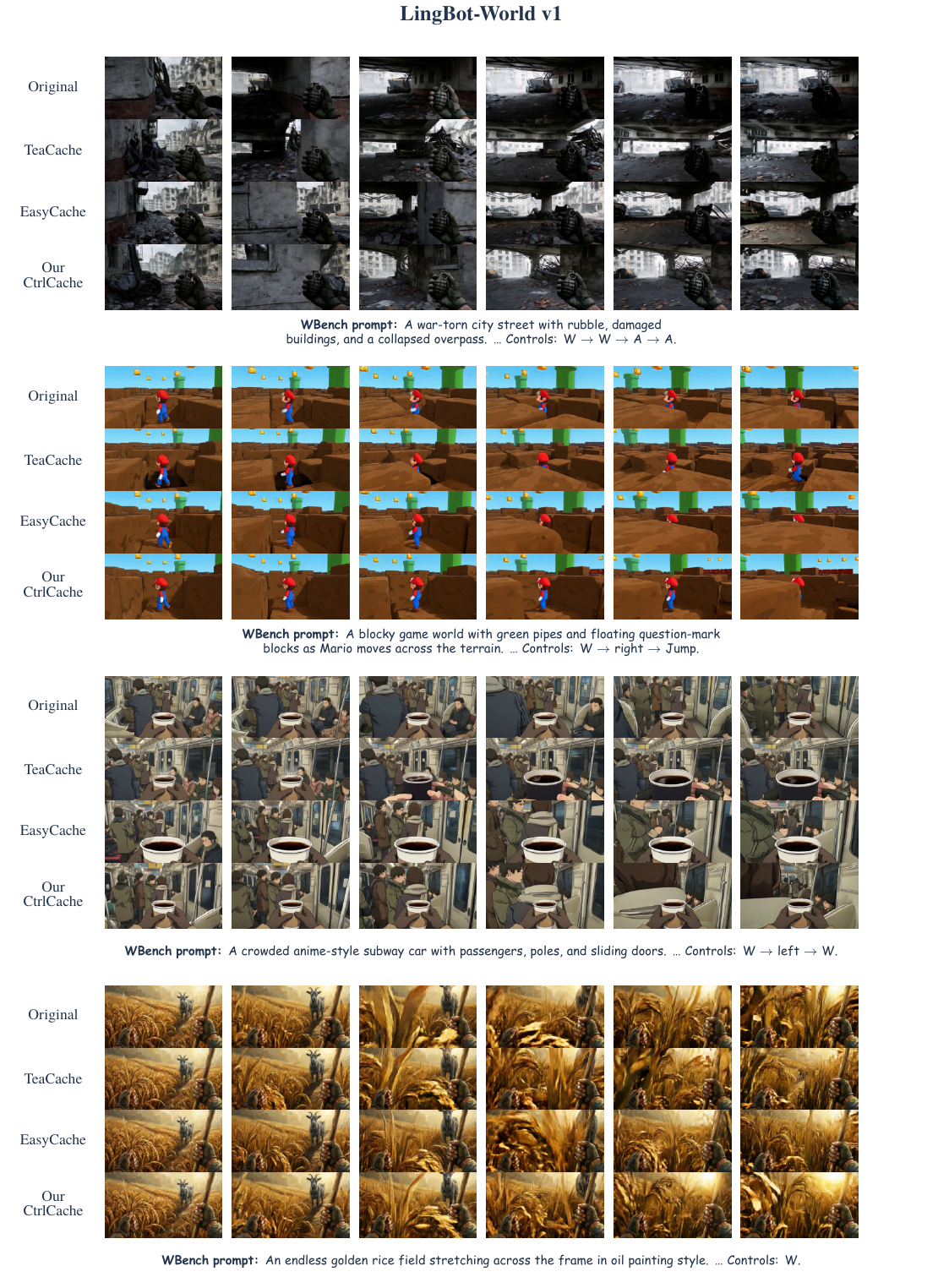}}
\end{figure}

\clearpage
\begin{figure}[p]
  \centering
  \makebox[\textwidth][c]{%
    \includegraphics[width=1.1\textwidth]{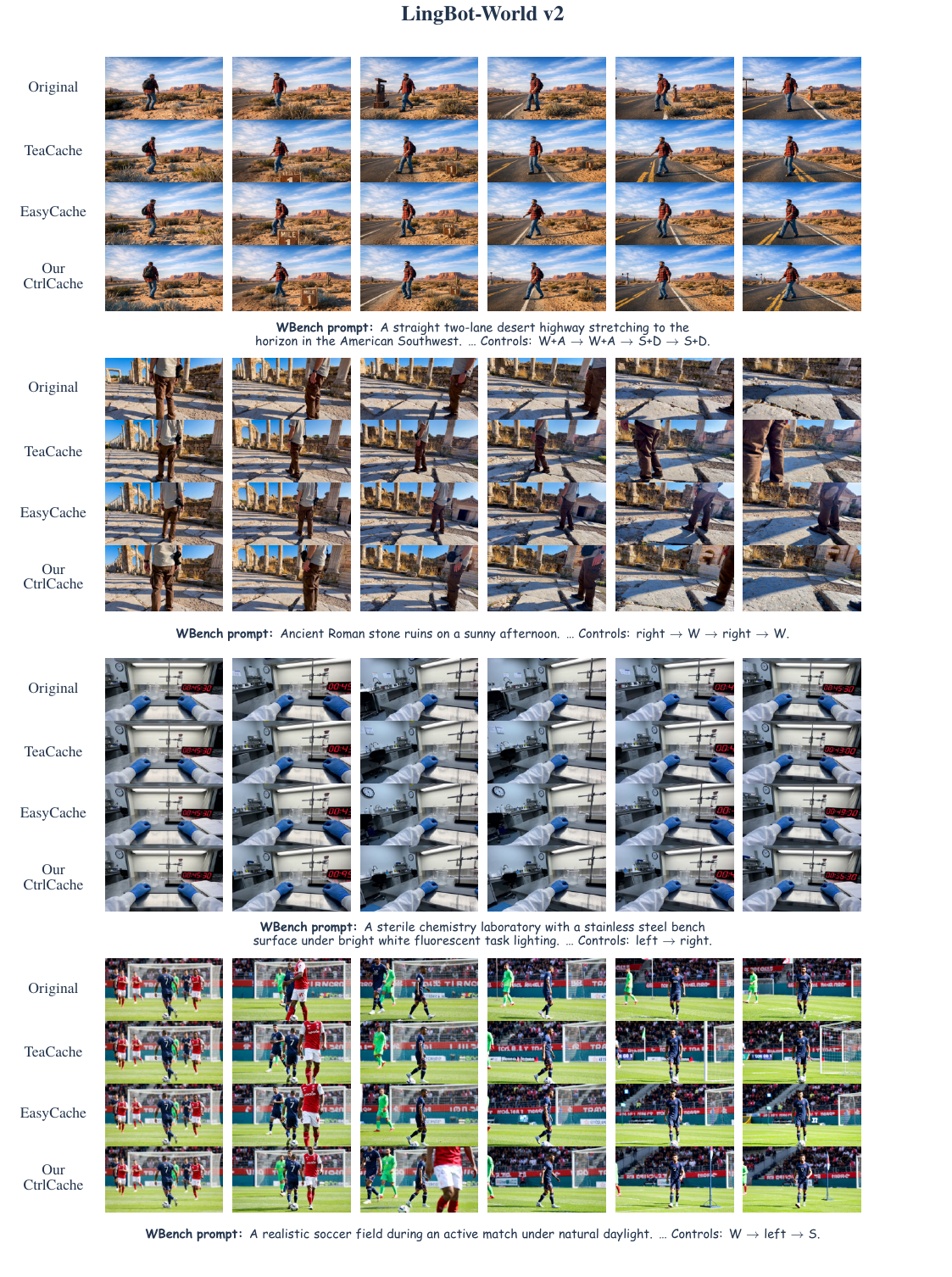}}
  \caption{Additional qualitative comparisons on Matrix-Game~2.0,
  LingBot-World v1, and LingBot-World v2. Each comparison uses the same
  initial condition and WBench control sequence across methods; the
  corresponding prompts and controls are shown with the frames. Rows
  correspond to Original, TeaCache, EasyCache, and Our CtrlCache.}
  \label{fig:supp_qualitative}
\end{figure}

\end{document}